# Graphical Models for Bandit Problems

Kareem Amin      Michael Kearns      Umar Syed
Department of Computer and Information Science
University of Pennsylvania
{akareem,mkearns,usyed}@cis.upenn.edu

## Abstract

We introduce a rich class of graphical models for multi-armed bandit problems that permit both the state or context space and the action space to be very large, yet succinctly specify the payoffs for any context-action pair. Our main result is an algorithm for such models whose regret is bounded by the number of parameters and whose running time depends only on the treewidth of the graph substructure induced by the action space.

## 1 Introduction

We introduce a rich class of graphical models for multi-armed bandit (MAB) problems that permit both the state or context space and the action space to be very large, yet succinctly specify the payoffs for any context-action pair. In our models there may be many context state variables, whose values are chosen exogenously by Nature, as well as many action variables, whose values must be chosen by an MAB algorithm. Only when all context and action variables are assigned values is an overall payoff observed. Thus each setting of the context variables yields a distinct but related bandit problem in which the number of actions scales exponentially.

Settings where the number of contexts and actions are both large are becoming common in applied settings such as sponsored search, where the number of possible user queries is effectively unbounded, and on many frequent queries the number of possible ads to show may also be large. Similarly, in many problems in quantitative trading, the number of ways one might break a trade up over multiple exchanges (the action space) is large, but might also depend on many conditioning variables such as market volatility (the context space). Recent lines of research have considered large parameterized context spaces and large parameterized action spaces separately (see Related Work below); here we are interested in models and algorithms for their simultaneous presence.

We consider a setting in which an MAB algorithm knows the coarse underlying graphical or dependency structure of the model, but not the actual parameters. Our main result is a no-regret algorithm for solving graphical MAB problems that is computationally efficient for a wide class of natural graphical structures. By no-regret we mean that the total regret of our algorithm is bounded by the number of parameters of the model, and thus is generally greatly sublinear in the number of contexts and actions. Interestingly, the running time of our algorithm depends only on properties of the substructure of the graphical model induced by the action variables — the dependencies between context variables, or between context and action variables, may be arbitrarily complex. If the treewidth of the action subgraph is constant, our algorithm runs in polynomial time.

The algorithm itself combines distributed dynamic programming — where the main challenge we face compared to standard such computations is the lack of any payoff until the entire model is instantiated (no local observations) — with the fact that our models admit linearization via a certain natural vector space of coefficients, which permits the application of recent "Knows What It Knows" (KWIK) algorithms for noisy linear regression. At a high level, for each context chosen by Nature, either our algorithm succeeds in choosing a reward-maximizing action (an exploitation step), or it effectively discovers another basis vector in a low-dimensional linear system determined by the graphical model (an exploration step).

The regret of the algorithm depends on the rank of this underlying vector space, which is always bounded by the number of parameters but may be smaller. It is a feature of our algorithm that no distributional assumptions are needed on the sequence of contexts chosen, which may be arbitrary. However, in these case that a distribution is present, the effective rank and thus our regret may be smaller still.

## 2 Related Work

Many authors have studied bandit problems where the number of states or *contexts* (each of which indexes a separate but perhaps related bandit problem) is allowed to be large (Auer et al., 2002; Langford and Zhang, 2007; Beygelzimer et al., 2011), while the number of actions, or the size of the function class to which the expected payoff function belongs, are assumed to be small (i.e. the sample and computational complexity of these algorithms are allowed to depend linearly on either of these quantities). In contrast, our results will consider both expected payoff function classes that are infinite, as well as context and action spaces that are assumed to be very large and thus call for sublinear complexity.

As we will demonstrate in Section 6, the setting we consider can be thought of as a linearly parameterized bandit problem. Such models associate each action $\mathbf{x}$ with a feature vector $\phi(\mathbf{x})$, and the expected payoff for taking that action is given by $\phi(\mathbf{x}) \cdot \mathbf{w}$, where $\mathbf{w}$ is an unknown weight vector. The computational complexity of most existing algorithms is nevertheless linear in the number of actions (Abe et al., 2003; Auer, 2002; Li et al., 2010). Furthermore, rather than being specified explicitly, the linear space in which our parameterizations lie are given by the underlying graphical or locality structure of the model.

More recent work on metric space contextual bandits (Slivkins, 2011; Lu et al., 2010) assumes that both action and context spaces are infinite, and that the expected payoff function comes from a large function class, namely the set of all Lipschitz continuous functions. The regret bounds given by both (Slivkins, 2011) and (Lu et al., 2010) are $\widetilde{O}(T^{(1+d)/(2+d)})$ where $d$ is some dimensionality constant dependent on the action and context spaces. In contrast, we make a different set of assumptions, namely that dependencies between actions admit a certain graphical representation, and provide regret bounds that are $\widetilde{O}(T^{2/3})$.

## 3 Preliminaries

We begin by assuming that both actions and contexts are represented by vectors of discrete variables, and that there is an unknown function which maps assignments to these variables to a payoff. The graphical assumption of our model will come when we assume how these variables interact.

Let $F : \prod_{i \in V} X_i \to [0,1]$ be the unknown *expected payoff function*, where each $X_i$ is the set of possible *values* for *variable* $i \in V$, with $|X_i| \geq 2$. The set of variables is partitioned into a subset $A \subseteq V$ of *action variables* and a subset $C \subseteq V$ (disjoint from $A$) of *context variables*, with $V = A \cup C$. Let $m = \max_i |X_i|$ and $n = |V|$.

For any subset of variables $S \subseteq V$, let $\mathbf{x}_S \in \prod_{i \in S} X_i$ denote a *joint assignment* of values to variables $S$. For shorthand, we write $\mathbf{X}_S = \prod_{i \in S} X_i$ to denote the set of all possible joint assignments to variables $S$. When $S = V$, we typically drop the subscript and write $\mathbf{x}$ and $\mathbf{X}$ instead of $\mathbf{x}_V$ and $\mathbf{X}_V$, and call $\mathbf{x}$ a *complete joint assignment*. We also abbreviate $\mathbf{x}_{\{i\}}$ as $\mathbf{x}_i$, when referring to the assignment of a single variable. If $S = A$ we call $\mathbf{x}_S$ a *joint action*, and if $S = C$ we call $\mathbf{x}_S$ a *joint context*.

### 3.1 Learning Protocol

Learning in our model proceeds as a series of trials. On each trial, Nature determines the current state or context, and our algorithm must choose values for the action variables in order to complete the joint assignment. Only then is a reward obtained, which depends on both the context and action.

In each round $t = 1, \ldots, T$:

1. Nature chooses an arbitrary joint context assignment $\mathbf{x}_C^t$, which is observed by the learning algorithm.

2. The learning algorithm chooses a joint action assignment $\mathbf{x}_A^t$.

3. The learning algorithm receives an independent random payoff $f^t \in [0,1]$ with expectation $F(\mathbf{x}_A^t, \mathbf{x}_C^t)$.

The *regret* after $T$ rounds is

$$R(T) \triangleq E\left[\sum_{t=1}^T \max_{\mathbf{x}_A} F(\mathbf{x}_A, \mathbf{x}_C^t) - F(\mathbf{x}_A^t, \mathbf{x}_C^t)\right]$$

where the expectation is with respect to randomness in the payoffs and the algorithm.

Note that in the learning protocol above, Nature chooses the context assignments as an arbitrary sequence; our main results hold for this rather strong setting. However, in Section 7.1, we also consider the special case in which each joint context $\mathbf{x}_C^t$ is drawn from a fixed distribution $\mathcal{D}$ (which is also the assumption of much of the previous research on contextual bandits), where better bounds may be obtained.

### 3.2 Assumption on Payoff Function

The main assumption that we leverage in this paper is that, while $F$ is unknown, we know that certain sets of variables may interact with each other to affect the payoff, while other sets may not. This is made precise in Assumption 1 below.

**Assumption 1** (Payoff Decomposition). *We are given a collection of variable subsets $\mathcal{P} \subseteq 2^V$ such that the unknown expected payoff function $F$ has the form*

$$F(\mathbf{x}) = \sum_{P \in \mathcal{P}} f_P(\mathbf{x}_P)$$

*where each unknown function $f_P : \mathbf{X}_P \to \mathbb{R}$ is called a* potential function.

We emphasize that the potential functions are unknown and arbitrary. What is known is that $F$ admits such a decomposition; more precisely, $\mathcal{P}$ is known.

Note that Assumption 1 is without loss of generality, since we can always take $\mathcal{P} = \{V\}$ and $f_V = F$. However, we are primarily interested in settings where $F$ decomposes into much simpler potential functions. If $|P| \leq k$ for all $P \in \mathcal{P}$ then we say the potential functions are $k$-ary.

Also note that Assumption 1 is very similar to the kinds of assumptions often made for tractable approximate inference (such as in a Markov random field), where a complex multivariate distribution is assumed to factorize into a product of several potential functions, each of which depend on only a subset of the variables. We next elaborate on the graph-theoretic aspects of our model.

### 3.3 Interaction Graphs

In the rest of the paper, we will use structural properties of the expected payoff function $F$ to bound the regret and running time of our graphical bandit algorithm. In particular, our results will depend on properties of the *interaction graph $G$* of the expected payoff function $F$. Let $G = (V, E)$ be a graph on variables $V$ with edges $E$ defined as follows: For any pair of variables $i, i' \in V$ we have edge $\{i, i'\} \in E$ if and only if there exists $P \in \mathcal{P}$ such that $i, i' \in P$, where $\mathcal{P}$ was defined in Assumption 1. In other words, we join $i$ and $i'$ by an edge if and only if there is some potential function $f_P$ that depends jointly on the variables $i$ and $i'$. The *action subgraph* $G_A = (A, E_A)$ is the subgraph of $G$ containing only the action variables $A \subseteq V$ and the edges between them $E_A \subseteq E$.

Note that the relationship between the interaction graph $G$ and an expected payoff function $F$ is essentially analogous to the relationship between a graphical model and a distribution whose independence structure it represents. The absence of an edge between two variables in a graphical model indicates that the variables are *probabilistically independent* in the distribution (conditioned on the remaining variables), while the absence of an edge between two variables in an interaction graph indicates that the two variables are *separable* in the expected payoff function. A sparse graphical model leads to computationally tractable inference algorithms, and we will shortly see that a sparse interaction graphs (more precisely, sparse action subgraphs) lead to computationally tractable no-regret algorithms for a graphical bandit problem.

To illustrate the role that the interaction graph plays in our approach, let us consider a simple contextual bandit problem that models a restricted version of sponsored web search. One of the most important categories of search queries in sponsored web search are those related to shopping, so we consider the task of serving ads in response to the search queries from users interested in purchasing airline tickets. Each query specifies the values of two context variables: the origin city $\mathbf{x}_{\text{origin}}$ and the destination city $\mathbf{x}_{\text{dest}}$. Each ad specifies the values of four action variables: the origin city $\mathbf{y}_{\text{origin}}$, the destination city $\mathbf{y}_{\text{dest}}$, the cost of the flight $\mathbf{y}_{\text{cost}}$ (e.g., 'cheap', 'first class'), and the brand of any accompanying hotel promotion $\mathbf{y}_{\text{hotel}}$ (e.g., 'none', 'Hyatt', 'Days Inn').[1]

The unknown expected payoff function $F$ specifies the clickthrough rate of each query/ad pair. Without further assumptions, $F$ may depend on all six variables jointly, and thus the interaction graph will be the fully-connected graph on six vertices. However, we can leverage our knowledge of the domain to decompose the expected payoff function and thus simplify the interaction graph. For any query/ad pair, we assume that the cities in the query affect the clickthrough rate only via their similarity, or lack thereof, with the cities contained in the ad. Also, we assume that the effect of an advertised hotel on the clickthrough rate is independent of the origin city, since presumably the hotel is located at the destination.

Leveraging the preceding assumptions, we assert that the unknown expected payoff function $F$ has the form

$$F(\mathbf{x}) = f_1(\mathbf{y}_{\text{origin}}, \mathbf{y}_{\text{cost}}, \mathbf{y}_{\text{dest}}) + f_2(\mathbf{y}_{\text{hotel}}, \mathbf{y}_{\text{cost}}, \mathbf{y}_{\text{dest}}) \\ + f_3(\mathbf{y}_{\text{origin}}, \mathbf{x}_{\text{origin}}) + f_4(\mathbf{y}_{\text{dest}}, \mathbf{x}_{\text{dest}})$$

(where, for notational brevity, we have subscripted the potential functions with numbers instead of subsets of variables). The interaction graph corresponding to this family of expected payoff functions is given in Figure 1, with context variables denoted by circles in the graph, and action variables by squares.

---

[1] For clarity, in this example we allow an action and context variable to share an index, and denote the action variable by $\mathbf{y}$ instead of $\mathbf{x}$.

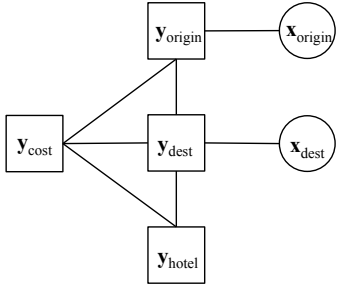

Figure 1: *Interaction graph for the simple sponsored search example described in Section 3. Context variables, which are related to the query, are denoted by circles. Action variables, which are related to the ad, are denoted by squares.*

The graph in Figure 1 is simpler than the fully-connected graph in at least two quantifiable ways: It has smaller maximum degree, and the treewidth of the subgraph of action variables is smaller. The regret and running time of the contextual bandit algorithm we present in Sections 4-7 will be bounded in terms of these two properties of the interaction graph.

We stress again that the *values* of the potential functions are not specified, and will be learned online by our algorithm. In this example, our algorithm might learn, for instance, that 'JFK' is a good substitute for 'New York', and that the users traveling to Las Vegas are more likely to click on ads that also offer rooms in luxury hotels.

Also, note that not all the independencies encoded by the expected payoff decomposition are preserved when forming the interaction graph. In the example above, the interaction graph would have been identical even if the potential functions $f_1$ and $f_2$ were merged into a single potential function. In this kind of situation, our algorithm implicitly uses an expected payoff decomposition in which each potential function corresponds to a maximal clique in the interaction graph. Certain probabilistic graphical models, such as Markov random fields, share a similar property.

## 4 Algorithm Overview

We present the details of our graphical bandit algorithm over the next three sections, and give a high-level overview in this section. Our approach will be to divide the graphical bandit problem into two subproblems — one essentially dealing with exploitation, and the other with exploration — and then compose the solutions to those subproblems into a no-regret algorithm.

In Section 5, we describe the `BestAct` algorithm, which, for any given joint context $\mathbf{x}_C$, computes an $\epsilon$-optimal joint action $\mathbf{x}_A^\epsilon$. The `BestAct` algorithm assumes access to an $\epsilon$-good approximation $F^\epsilon(\cdot, \mathbf{x}_C)$ of the expected payoff $F(\cdot, \mathbf{x}_C)$, and when computing $\mathbf{x}_A^\epsilon$ for given joint context $\mathbf{x}_C$ the `BestAct` algorithm makes many calls to the oracle $F^\epsilon(\cdot, \mathbf{x}_C)$. Note that `BestAct` cannot simply issue these queries to Nature instead, since it has no way of fixing the joint context $\mathbf{x}_C$ over several rounds. The running time of `BestAct` is polynomially bounded if the treewidth of the action subgraph $G_A$ is constant. `BestAct` is related to existing dynamic programming algorithms for finding the maximum *a posteriori* (MAP) assignment in a graphical model, but unlike that setting must overcome the additional challenge of receiving only *global* payoffs and no direct local observations or information.

In Section 6, we describe the `PayEst` algorithm, which implements the oracle required by `BestAct`. With high probability, whenever `PayEst` receives a complete joint assignment $(\mathbf{x}_A, \mathbf{x}_C)$, either it outputs the value of $F^\epsilon(\mathbf{x}_A, \mathbf{x}_C)$, or it outputs a special symbol $\perp$. `PayEst` is an instance of a "Knows What It Knows" (KWIK) algorithm (Li et al., 2008), and the number of times `PayEst` outputs $\perp$ is polynomially-bounded if the potential functions are all $k$-ary for a constant $k$. The bound is a consequence of the fact that $F$ can be written as a linear function, and each $\perp$ output increments the dimension of a certain linear subspace. Again, note that `PayEst` cannot be as simple as repeatedly playing the complete joint assignment $(\mathbf{x}_A, \mathbf{x}_C)$ and averaging the observed payoffs, since our algorithm can only specify the joint action $\mathbf{x}_A$ in each round, and has no way of fixing the joint context $\mathbf{x}_C$.

In Section 7, we put `BestAct` and `PayEst` together to form `GraphicalBandits`, an algorithm for graphical bandit problems, which in each round $t$ runs `BestAct` for the current context $\mathbf{x}_C^t$, and uses `PayEst` to implement the oracle required by `BestAct`. The main difficulty in integrating the two algorithms is that sometimes `PayEst` outputs $\perp$ instead of the value of the oracle $F^\epsilon(\cdot, \mathbf{x}_C^t)$. However, whenever this happens, `BestAct` will provide feedback that causes `PayEst` to make measurable exploration progress, improving its ability to provide accurate payoff estimates in subsequent rounds.

## 5 Best Action Subroutine

We will now describe `BestAct`, an algorithm that efficiently computes, for a given joint context $\mathbf{x}_C$, an $\epsilon$-optimal joint action $\mathbf{x}_A^\epsilon$ satisfying $F(\mathbf{x}_A^\epsilon, \mathbf{x}_C) \geq$

$\max_{\mathbf{x}_A} F(\mathbf{x}_A, \mathbf{x}_C) - \epsilon$. The `BestAct` algorithm uses dynamic programming applied to the action subgraph $G_A$, and is similar to standard dynamic programming algorithms, such as the Viterbi algorithm, for computing the MAP assignment in a graphical model. Our setting is more challenging, however. For one, we do not have access to the individual potential functions $f_P$, but only to the global expected payoff function $F$ (assuming no noise). Furthermore, we do not control the context argument to $F$.

To illustrate these difficulties, suppose that the action subgraph $G_A$ is a tree, and that we wish to run a standard dynamic programming algorithm which, fixing an assignment for a variable's parent, attempts to find the best assignment for the subtree rooted at that variable, under a joint context $\mathbf{x}_C$. That algorithm would require several queries to $F(\cdot, \mathbf{x}_C)$. However, once the first observation is received, the next context is not guaranteed to be the same, and the algorithm suffers regret for that round. Furthermore, as we have previously remarked, any observation $F(\mathbf{x}_A, \mathbf{x}_C)$, does not give us the values of the individual potentials $f_P(\mathbf{x}_P)$. Thus, to compare two joint assignments to a subtree in the algorithm sketched, we must be careful to fix the assignment to all variables outside the subtree in question. This becomes even more delicate when $G_A$ is not a tree.

In order to overcome these problems, we will assume oracle access to a function $F^\epsilon(\cdot, \mathbf{x}_C)$ for the given joint context $\mathbf{x}_C$ that satisfies $|F^\epsilon(\mathbf{x}_A, \mathbf{x}_C) - F(\mathbf{x}_A, \mathbf{x}_C)| \leq \epsilon$ for all joint actions $\mathbf{x}_A$. In Section 6 we describe the `PayEst` algorithm, which implements this oracle.

For ease of exposition, we only give a detailed algorithm and analysis for the case when $G_A$ is acyclic, although we have a related algorithm that can be applied when $G_A$ is an arbitrary graph. We state our results for arbitrary graphs, without proofs, at the end of this section.

Suppose that the action subgraph $G_A$ is a tree. (Note that this implies that each potential function $f_P$ depends jointly on at most two variables). Let $r$ be an arbitrarily chosen root for $G_A$. For any $a \in A$, let $T(a)$ denote the vertices of the subtree rooted at $a$. Let $ch(a)$ denote the set of vertices which are children of $a$, and for $a \neq r$, let $pa(a)$ denote the parent of $a$. Also for $a \neq r$, let $R(a) = A \setminus (T(a) \cup \{pa(a)\})$ be remaining vertices that are neither $pa(a)$ nor belong to $T(a)$.

If $S_1, \ldots, S_k \in 2^V$ are mutually disjoint subsets of variables such that $S = S_1 \cup \cdots \cup S_k$, we write a joint assignment $\mathbf{x}_S$ as $\mathbf{x}_S = (\mathbf{x}_{S_1}, \ldots, \mathbf{x}_{S_k})$.

For $a \neq r$, we would like to compute the best joint assignment to the variables of $T(a)$, having fixed an assignment for $pa(a)$. We will denote this best joint assignment by $[a, \mathbf{x}_{pa(a)}]^*$. For any leaf $a$ and assignment $\mathbf{x}_{pa(a)}$,

$$[a, \mathbf{x}_{pa(a)}]^* = \arg\max_{\mathbf{x}_a} F(\mathbf{x}_a, \mathbf{x}_{pa(a)}, \mathbf{x}'_{R(a)}, \mathbf{x}_C)$$

where $\mathbf{x}'_{R(a)}$ is a fixed, but arbitrary joint assignment to the vertices in $R(a)$. For a fixed choice of $\mathbf{x}_{pa(a)}$, $[a, \mathbf{x}_{pa(a)}]^*$ is the same regardless of how $\mathbf{x}'_{R(a)}$ is chosen since, by assumption, we can write $F(\mathbf{x}_a, \mathbf{x}_{pa(a)}, \mathbf{x}'_{R(a)}, \mathbf{x}_C) = f_{\{pa(a),a\}}(\mathbf{x}_{pa(a)}, \mathbf{x}_a, \mathbf{x}_C) + \sum_{e \in E_A \setminus \{(p(a),a)\}} f_e(\mathbf{x}_p, \mathbf{x}_C)$, where second term is a constant with respect to $\mathbf{x}_a$. If we had access to $F(\cdot, \mathbf{x}_C)$, $[a, \mathbf{x}_{pa(a)}]^*$ can be efficiently computed.

In general, for $a \neq r$ not necessarily a leaf, and $ch(a) = \{a_1, \ldots, a_d\}$, we can write:

$$\mathbf{x}^*_{a, \mathbf{x}_{pa(a)}} = \arg\max_{\mathbf{x}_a} F(\mathbf{x}_a, [a_1, \mathbf{x}_a]^*, \ldots, [a_d, \mathbf{x}_a]^*, \mathbf{x}_{pa(a)}, \mathbf{x}'_{R(a)}, \mathbf{x}_C)$$

and

$$[a, \mathbf{x}_{pa(a)}]^* = (\mathbf{x}^*_{a, \mathbf{x}_{pa(a)}}, [a_1, \mathbf{x}^*_{a, \mathbf{x}_{pa(a)}}]^*, \ldots, [a_d, \mathbf{x}^*_{a, \mathbf{x}_{pa(a)}}]^*)$$

This motivates an algorithm, wherein the values $[a, \mathbf{x}_{pa(a)}]^*$ are computed for each $a$ and all choices $\mathbf{x}_{pa(a)}$. This can be done efficiently by taking $a$ in postfix order (children before parents).

However, we only have access to the approximate oracle $F^\epsilon(\cdot, \mathbf{x}_C)$. Therefore, we will instead consider:

$$\mathbf{x}^{*,\epsilon}_{a, \mathbf{x}_{pa(a)}} = \arg\max_{\mathbf{x}_a} F^\epsilon(\mathbf{x}_a, [a_1, \mathbf{x}_a]^\epsilon, \ldots, [a_d, \mathbf{x}_a]^\epsilon, \mathbf{x}_{pa(a)}, \mathbf{x}'_{R(a)}, \mathbf{x}_C)$$

and

$$[a, \mathbf{x}_{pa(a)}]^\epsilon = (\mathbf{x}^{*,\epsilon}_{a, \mathbf{x}_{pa(a)}}, [a_1, \mathbf{x}^{*,\epsilon}_{a, \mathbf{x}_{pa(a)}}]^\epsilon, \ldots, [a_d, \mathbf{x}^{*,\epsilon}_{a, \mathbf{x}_{pa(a)}}]^\epsilon)$$

where each $\mathbf{x}'_{R(a)}$ is selected arbitrarily. We now argue that the $[a, \mathbf{x}_{pa(a)}]^\epsilon$ values, computed using the function $F^\epsilon(\cdot, \mathbf{x}_C)$, are good values with respect to the true expected payoff function $F(\cdot, \mathbf{x}_C)$.

**Theorem 1.** *For any action variable $a \in A$, and assignments $\mathbf{x}_{pa(a)}, \mathbf{x}'_{R(a)}$,*

$$F(\mathbf{x}_{pa(a)}, [a, \mathbf{x}_{pa(a)}]^*, \mathbf{x}'_{R(a)}, \mathbf{x}_C)$$
$$\leq F(\mathbf{x}_{pa(a)}, [a, \mathbf{x}_{pa(a)}]^\epsilon, \mathbf{x}'_{R(a)}, \mathbf{x}_C) + 2|T(a)|\epsilon$$

*Proof.* Fix a choice of $\mathbf{x}'_{R(a)}$ and $\mathbf{x}_{pa(a)}$. Let $a$ be a leaf. We have that:

$$F([a, \mathbf{x}_{pa(a)}]^*, \mathbf{x}_{pa(a)}, \mathbf{x}'_{R(a)}, \mathbf{x}_C) - \epsilon$$
$$\leq F^\epsilon([a, \mathbf{x}_{pa(a)}]^*, \mathbf{x}_{pa(a)}, \mathbf{x}'_{R(a)}, \mathbf{x}_C)$$
$$\leq F^\epsilon([a, \mathbf{x}_{pa(a)}]^\epsilon, \mathbf{x}_{pa(a)}, \mathbf{x}'_{R(a)}, \mathbf{x}_C)$$
$$\leq F([a, \mathbf{x}_{pa(a)}]^\epsilon, \mathbf{x}_{pa(a)}, \mathbf{x}'_{R(a)}, \mathbf{x}_C) + \epsilon$$

Rearranging establishes the inequality. Now suppose that the claim is true for each $a' \in ch(a) = \{a_1, \ldots, a_d\}$ for some variable $a$ and assignment $\mathbf{x}_{p(a)}$.

$$F([a, \mathbf{x}_{pa(a)}]^*, \mathbf{x}_{pa(a)}, \mathbf{x}'_{R(a)}, \mathbf{x}_C)$$
$$= F(x^*_{a,pa(a)}, [a_1, x^*_{a,pa(a)}]^*, \ldots,$$
$$[a_d, x^*_{a,pa(a)}]^*, \mathbf{x}_{pa(a)}, \mathbf{x}'_{R(a)}, \mathbf{x}_C)$$
$$\leq F(x^*_{a,pa(a)}, [a_1, x^*_{a,pa(a)}]^\epsilon, \ldots,$$
$$[a_d, x^*_{a,pa(a)}]^\epsilon, \mathbf{x}_{pa(a)}, \mathbf{x}'_{R(a)}, \mathbf{x}_C) + 2\sum_{i=1}^{d} |T(a_i)|\epsilon,$$

where the inequality is by invoking the induction hypothesis for each $[a_i, x^*_{a,pa(a)}]^*$. Similar reasoning as the base case gives us that this last expression is less than or equal to

$$F(x^{*,\epsilon}_{a,pa(a)}, [a_1, x^{*,\epsilon}_{a,pa(a)}]^\epsilon, \ldots,$$
$$[a_d, x^{*,\epsilon}_{a,pa(a)}]^\epsilon, \mathbf{x}_{pa(a)}, \mathbf{x}'_{R(a)}, \mathbf{x}_C) + 2|T(a)|\epsilon,$$

completing the proof. □

Imagining a "super-root" $r'$ such that $r' = pa(r)$ and $f_{\{r',r\}} \equiv 0$, selecting $a = r$, and selecting $\mathbf{x}_{r'}$ arbitrarily in Theorem 1, gives us an algorithm which will compute an $2|A|\epsilon$-optimal joint action, given access to $F^\epsilon$. We shall refer to this algorithm as BestAct.

**Theorem 2.** *Suppose the action subgraph $G_A$ is a tree. Given access to oracle $F^\epsilon(\cdot, \mathbf{x}_C)$, the BestAct algorithm computes a $2|A|\epsilon$-optimal joint action for joint context $\mathbf{x}_C$ in $O(m^2|E_A|)$ time.*

*Proof.* Fix an arbitrary variable $a$. Given $[a_i, \mathbf{x}_{pa(a_i)}]^\epsilon$ for each $a_i \in ch(a)$, $[a, \mathbf{x}_{pa(a)}]^\epsilon$ can be computed in $O(m)$ time by definition. Thus, computing $[r, \mathbf{x}_{r'}]^\epsilon$ can be done in $O(m^2|A|)$ time. □

Notice that any acyclic forest $G_A$ can be handled by running the tree algorithm on each connected component. Suppose now that the action subgraph $G_A$ is arbitrary, but admits a tree decomposition $\mathcal{T} = (\mathcal{A}, \mathcal{E})$ where $S \in \mathcal{A}$ is a subset of $A$. Let $w = \max_{S \in \mathcal{A}} |S|$, the treewidth of $G_A$.

**Theorem 3.** *Let $\mathcal{T} = (\mathcal{A}, \mathcal{E})$ be a tree decomposition of action subgraph $G_A$ with treewidth $w$. Given access to oracle $F^\epsilon(\cdot, \mathbf{x}_C)$, the BestAct algorithm can be generalized to compute a $2|A|\epsilon$-optimal joint action for joint context $\mathbf{x}_C$ in $O(m^{2w}|\mathcal{E}|)$ time.*

*Proof idea.* The approach used when $G_A$ is a tree can be generalized to run on the tree decomposition of an arbitrary action subgraph. The development is very similar to then recursive argument given in this section, and the details are omitted due to space constraints. □

Note that in order for the generalized version of BestAct to be computationally efficient, natural but nontrivial restrictions on the action subgraph $G_A$ are required (namely, small treewidth). It can be shown that some restrictions are inevitable. For example, the energy $F$ of a 3-D Ising model from statistical physics can be phrased as the sum of binary potential functions. However, even if the behavior of each potential function $f_P$ is known (i.e. there is no implicit learning problem), and there are no contexts, it is still NP-hard to select the action (i.e. spins on the variables of the Ising graph) maximizing the energy function $F$ (Barahona, 1982).

## 6 Payoff Estimator Subroutine

In this section we present the PayEst algorithm, which implements the $\epsilon$-good oracle $F^\epsilon(\mathbf{x}_A, \mathbf{x}_C)$ required by the BestAct algorithm described in Section 5. Before presenting PayEst, we give in Section 6.1 a precise definition of the problem that it is designed to solve. In Section 6.2 we give a simple solution for the special case where the observed payoffs are deterministic, which suffices to convey the main ideas of our approach. The solution to the general problem is given in Section 6.3, which will be an instance of a "Knows What It Knows" (KWIK) algorithm (Li et al., 2008).

### 6.1 Problem Statement

Let us review the details of how the BestAct algorithm uses its oracle. On several time steps[2] $s = 1, \ldots, S$ BestAct specifies a complete joint assignment $\mathbf{x}^s = (\mathbf{x}^s_A, \mathbf{x}^s_C)$ and requests the value $F^\epsilon(\mathbf{x}^s)$ from the oracle. Ideally, we would like PayEst to unerringly supply these values to BestAct. However, since the true expected payoff function $F$ is unknown, this will be impossible in general. Instead, PayEst will be designed for the following learning protocol: On each time step $s$ the algorithm must *either* output the value $\hat{f}^s = F^\epsilon(\mathbf{x}^s)$ *or* output a special symbol $\bot$ and be allowed to observe an independent random variable $f^s$ with mean $F(\mathbf{x}^s)$. As we shall see in Section 7, BestAct and PayEst can be integrated in a way that respects this protocol and also bounds the number of times that PayEst outputs $\bot$.

### 6.2 Deterministic Payoffs Setting

Let us first consider the special case where each payoff $f^s$ is in fact deterministic; that is, PayEst always observes $F(\mathbf{x}^s)$ directly. We give a simple algorithm for PayEst that is based on the idea that Assumption 1 allows us to express the expected payoff function $F$ in a compact *linearized* form. We now proceed to describe this linearization, and then give the algorithm.

---
[2]These time steps are *not* the rounds of the bandit problem; BestAct may call the oracle several times per round of the bandit problem.

For convenience, let $N = \sum_{P \in \mathcal{P}} \prod_{i \in P} |X_i|$, and define the *payoff vector* $\mathbf{f} \in \mathbb{R}^N$ as follows: Divide the $N$ components of $\mathbf{f}$ into $|\mathcal{P}|$ blocks, where each block corresponds to a potential function $f_P$. Within the block for potential function $f_P$, let there be one component $\ell$ corresponding to each of the $\prod_{i \in P} |X_i|$ possible joint assignments $\mathbf{x}_P$, and set the $\ell$th component of $\mathbf{f}$ equal to $f_P(\mathbf{x}_P)$.

For any complete joint assignment $\mathbf{x}$, let $\mathbf{v}(\mathbf{x}) \in \{0,1\}^N$ be a binary *coefficient vector* defined as follows: If $\ell$ is the component of $\mathbf{f}$ corresponding to potential function $f_P$ and joint assignment $\mathbf{x}'_P$ then the $\ell$th component of $\mathbf{v}(\mathbf{x})$ equals 1 if and only if $\mathbf{x}_P = \mathbf{x}'_P$.

For an illustration of a payoff vector and coefficient vector, consider an expected payoff function $F$ that depends on three variables and decomposes into two potential functions. Let $F(x_1, x_2, x_3) = f_{\{1,2\}}(x_1, x_2) + f_{\{2,3\}}(x_2, x_3)$, and suppose that the first potential adds its inputs, while the second potential multiplies them, i.e., $f_{\{1,2\}}(x_1, x_2) = x_1 + x_2$ and $f_{\{2,3\}}(x_2, x_3) = x_2 x_3$. If each variable $x_i$ takes its values from the set $\{a, b\}$, then the payoff vector for the function $F$ can be written $\mathbf{f} = (a+a, a+b, b+a, b+b, aa, ab, ba, bb)$ and the coefficient vector corresponding to, say, the complete joint assignment $\mathbf{x} = (x_1, x_2, x_3) = (a, b, a)$ is $\mathbf{v}(\mathbf{x}) = (0, 1, 0, 0, 0, 0, 1, 0)$. Most importantly, note that in general, the definitions of $\mathbf{f}$ and $\mathbf{v}(\mathbf{x})$, together with Assumption 1, imply that the expected payoff function has the linear form $F(\mathbf{x}) = \mathbf{f} \cdot \mathbf{v}(\mathbf{x})$.

Now a very natural PayEst algorithm presents itself: On each time step $s$, if there exists a linear combination $\alpha_1, \ldots, \alpha_{s-1}$ such that $\mathbf{v}(\mathbf{x}^s) = \sum_{s'=1}^{s-1} \alpha_{s'} \mathbf{v}(\mathbf{x}^{s'})$ — in other words, if $\mathbf{v}(\mathbf{x}^s)$ is in the linear span of previous coefficient vectors — then output the estimate $\hat{f}^s = \sum_{s'=1}^{s-1} \alpha_{s'} f^{s'}$, and otherwise output $\perp$. Clearly, because the expected payoff function $F(\mathbf{x})$ is a linear function of the coefficient vector $\mathbf{v}(\mathbf{x})$ and the observed payoffs are deterministic, we have $\hat{f}^s = F(\mathbf{x}^s)$ for all $s$. Also, since each coefficient vector is in $\mathbb{R}^N$, and there is no set of linearly independent vectors in $\mathbb{R}^N$ containing more than $N$ vectors, the number of observation time steps is at most $N$. Importantly, we can upper bound $N$ in terms of properties of the interaction graph $G$. Recall that $m = \max_i |X_i|$ and $n = |V|$, and suppose that each potential function $P \in \mathcal{P}$ is $k$-ary. We have $N = \sum_{P \in \mathcal{P}} \prod_{i \in P} |X_i| \leq |\mathcal{P}| m^k \leq \binom{n}{k} m^k \leq (mn)^k$. Thus, if we regard $k$ as a constant, the number of observation time steps is upper bounded by a polynomial.

In fact, we can further exploit the structure of the interaction graph $G$ to give a more refined upper bound than $N$. Let $M = \prod_{i \in V} |X_i|$, and define the *coefficient matrix* $\mathbf{M} \in \{0,1\}^{N \times M}$ as follows: For each of the $M$ possible complete joint assignments $\mathbf{x}$, the matrix $\mathbf{M}$ contains one column that equals the coefficient vector $\mathbf{v}(\mathbf{x})$. Since there is no set of linearly independent columns of $\mathbf{M}$ containing more than $\text{rank}(\mathbf{M})$ vectors, the number of observation time steps is at most $\text{rank}(\mathbf{M})$, which is at most $N$, but potentially much less than $N$. In Section 8, we give another result bounding $\text{rank}(\mathbf{M})$ in terms of properties of the interaction graph $G$.

### 6.3 Probabilistic Payoffs Setting

We now return to the general setting, so that each $f^s$ is no longer deterministic, but an independent random variable with expected value $F(\mathbf{x}^s)$. Thanks to the linearized representation of the expected payoff function $F$ described in the previous section, we can use a *KWIK linear regression algorithm* (Strehl and Littman, 2007) to implement PayEst. On each time step $s = 1, \ldots, S$ such an algorithm observes a feature vector $\boldsymbol{\phi}^s$ and does exactly one of the following: (1) outputs prediction $\hat{y}^s$, or (2) outputs $\perp$ and observes independent random variable $y^s \in [0,1]$ with expected value $\mathbf{w} \cdot \boldsymbol{\phi}^s$, where the weights $\mathbf{w}$ are unknown. In our case, $\hat{y}^s$ and $y^s$ are the predicted and observed payoffs $\hat{f}^s$ and $f^s$, the feature vector $\boldsymbol{\phi}^s$ is the coefficient vector $\mathbf{v}(\mathbf{x}^s)$, and the unknown weight vector $\mathbf{w}$ is the payoff vector $\mathbf{f}$.

There are several existing algorithms for KWIK linear regression. For example, Cesa-Bianchi et al. (2009) describe an algorithm for which the number of observation time steps is upper bounded by $O\left(\frac{d}{\epsilon^2} \log(\frac{S}{\epsilon \delta})\right)$, where $d$ is the dimension of the subspace containing the feature vectors. In our case, we have $d = \text{rank}(\mathbf{M})$, where $\mathbf{M}$ is the coefficient matrix. For concreteness, we will henceforth use this algorithm for PayEst.

## 7 Graphical Bandits Algorithm

In this section, we compose the algorithms from the previous two sections to form the GraphicalBandits algorithm, which is described in detail in Algorithm 1. In each round $t$, GraphicalBandits runs BestAct on the current joint context $\mathbf{x}_C^t$, and whenever BestAct asks the oracle for the value of $F^\epsilon(\mathbf{x}_A, \mathbf{x}_C^t)$ for some joint action $\mathbf{x}_A$, this request is passed on to PayEst as the coefficient vector $\mathbf{v}(\mathbf{x}_A, \mathbf{x}_C^t)$. If PayEst never returns $\perp$ for a given run of BestAct, then by the analysis in Sections 5 and 6, BestAct will return an $\epsilon$-optimal joint action $\mathbf{x}_A^\epsilon$ for joint context $\mathbf{x}_C^t$, which is then played in round $t$. However, if PayEst returns $\perp$ in response to some complete joint assignment

$(\mathbf{x}_A, \mathbf{x}_C^t)$, then `BestAct` is terminated immediately, the joint action $\mathbf{x}_A$ is played in round $t$, and the observed payoff $f^t$ is returned to `PayEst`. Since `PayEst` is a KWIK linear regression algorithm, this feedback is required in order to bound the number of times that `PayEst` outputs $\perp$.

---
**Algorithm 1** `GraphicalBandits`
---
1: **Given:** Subroutine `PayEst`, subroutine `BestAct`, parameters $\epsilon, \delta$.
2: Initialize `PayEst` with parameters $\epsilon, \delta$.
3: **for** each time step $t = 1, \ldots, T$ **do**
4:    Run `BestAct` on observed joint context $\mathbf{x}_C^t$.
5:    **while** `BestAct` is running **do**
6:      **if** `BestAct` asks for value of $F^\epsilon(\mathbf{x}_A, \mathbf{x}_C^t)$ **then**
7:        Input coeff. vector $\mathbf{v}(\mathbf{x}_A, \mathbf{x}_C^t)$ to `PayEst`.
8:        **if** `PayEst` returns $\perp$ **then**
9:          Play joint action $\mathbf{x}_A$, observe payoff $f^t$.
10:          Return $f^t$ to `PayEst`.
11:          Terminate `BestAct` early.
12:        **else if** `PayEst` returns $\hat{f}$ **then**
13:          Return $\hat{f}$ to `BestAct`.
14:        **end if**
15:      **end if**
16:    **end while**
17:    **if** `BestAct` was not terminated early **then**
18:      Play joint action $\mathbf{x}_A^\epsilon$ returned by `BestAct`.
19:    **end if**
20: **end for**

---

**Theorem 4.** *Let $d = \text{rank}(\mathbf{M})$ be the rank of the coefficient matrix, and let $\mathcal{T} = (\mathcal{A}, \mathcal{E})$ be the tree decomposition of the action subgraph $G_A$ with treewidth $w$. The regret $R(T)$ of `GraphicalBandits` after $T$ rounds is at most $R(T) \leq O\left(\frac{dw}{\epsilon^2} \log(Tm|\mathcal{E}|/\epsilon\delta) + 2|\mathcal{A}|\epsilon T + \delta T\right)$ and the computational complexity of each round is $O(m^{2w}|\mathcal{E}|)$.*

*Proof.* The per round computational complexity follows from Theorem 3.

Since `PayEst` is the KWIK linear regression algorithm due to Cesa-Bianchi et al. (2009), we have that with probability $1 - \delta$, every prediction $\hat{f}$ has error at most $\epsilon$, and $\perp$ is returned at most $K \leq O\left(\frac{d}{\epsilon^2} \log(\frac{S}{\epsilon\delta})\right)$ times, where $S$ is the number of times `BestAct` calls `PayEst`. Since `BestAct` call `PayEst` at most $O(m^{2w}|\mathcal{E}|)$ times per round, we have $S \leq O(Tm^{2w}|\mathcal{E}|)$. In the probability $\delta$ event that `PayEst` fails to meet its guarantee, we may suffer maximum regret on all $T$ rounds. Otherwise, we are guaranteed that `BestAct` completes successfully on all but $K$ rounds, and on the remaining rounds, by Theorem 3, a $2|\mathcal{A}|\epsilon$-optimal joint action is selected.

Thus, $R(T) \leq K + 2|\mathcal{A}|\epsilon T + \delta T \leq O\left(\frac{wd}{\epsilon^2} \log(Tm|\mathcal{E}|/\epsilon\delta) + 2|\mathcal{A}|\epsilon T + \delta T\right)$ □

If we tune $\epsilon$ and $\delta$ appropriately in terms of the number of rounds $T$, we obtain the following no-regret bound.

**Corollary 1.** *If `GraphicalBandits` is run with parameters $\epsilon = \delta = \frac{1}{T^{1/3}}$ then $R(T) \leq O\left(dw|\mathcal{A}|T^{2/3} \log(Tm|\mathcal{E}|)\right)$.*

In Corollary 1, the dependence of regret on the number of rounds is $\widetilde{O}(T^{2/3})$, while the regret bounds for many bandit algorithms are $\widetilde{O}(\sqrt{T})$, raising the question of whether our regret bound can be improved. The VE algorithm of Beygelzimer et al. (2011) is an *inefficient* (i.e., exponential-time) contextual bandit algorithm that has regret $\widetilde{O}(\sqrt{T})$. Whether there exists an polynomial-time algorithm that achieves $\widetilde{O}(\sqrt{T})$ regret is an open problem; Abernathy and Rakhlin (2009) supply a relevant discussion.

### 7.1 Distributional Assumptions

So far we have considered a setting in which contexts are chosen arbitrarily, and our regret bound in Theorem 4 depends on $d = \text{rank}(\mathbf{M})$. This is because, in the worst-case, each coefficient vector $\mathbf{v}(\mathbf{x}_A, \mathbf{x}_C)$ observed by `PayEst` will be linearly independent of all previously observed coefficient vectors. However, note that if the joint contexts are drawn from a distribution, then such a worst-case sequence may no longer be likely. Let $\mathcal{D}$ be an unknown *context distribution* on the joint contexts $\mathbf{x}_C$.

Fix a set $\mathbf{X}_C' \subseteq \mathbf{X}_C$, and let $\mathbf{M}(\neg \mathbf{X}_C')$ be the coefficient matrix restricted to joint contexts in $\mathbf{X}_C \setminus \mathbf{X}_C'$. In other words, the columns of $\mathbf{M}(\neg \mathbf{X}_C')$ are exactly $\{\mathbf{v}(\mathbf{x}_C, \mathbf{x}_A) \mid \mathbf{x}_A \in \mathbf{X}_A, \mathbf{x}_C \in \mathbf{X}_C \setminus \mathbf{X}_C'\}$. We can provide an alternate regret bound for the same algorithm by conceding full regret on rounds in which a context is chosen from $\mathbf{X}_C'$, and considering the algorithm's performance only on the remaining rounds. This gives us

$$R(T) \leq TP_{\mathbf{x}_C \sim \mathcal{D}}(\mathbf{x}_C \in \mathbf{X}_C') + \\ O\left(\text{rank}(\mathbf{M}(\neg \mathbf{X}_C'))w|\mathcal{A}|T^{2/3} \log(Tm|\mathcal{E}|)\right)$$

Thus if $\mathcal{D}$ places a large amount of its mass on joint contexts that generate a coefficient matrix with low rank, the bound above may be significantly better than our general bound. Note we can optimize this bound for rare contexts under the distribution $\mathcal{D}$ by minimizing this quantity over sets of joint contexts $\mathbf{X}_C'$.

## 8 Rank and Graph Structure

Let $\mathbf{M}_F$ be the coefficient matrix corresponding to an expected payoff function $F$, and $G_F = (V, E)$ be its interaction graph. In this section, we observe that we can use structural properties of $G_F$ to prove statements about rank($\mathbf{M}_F$), and therefore the regret of our algorithm. Consider an expected payoff function $F$ that is the sum of binary potential functions (i.e., $k$-ary with $k = 2$). In Section 6 we argued that rank($\mathbf{M}$) = $O(|V|^2)$. In this section we will argue that rank($\mathbf{M}$) = $\Omega(|V|)$.

**Theorem 5.** *For any $F$ with binary potentials, rank($\mathbf{M}_F$) = $\Omega(|V|)$.*

*Proof.* Suppose that the largest matching in $G_F$ contains at least $|V|/4$ edges. Since each potential function is binary, each edge in $G_F$ represents a potential function in the decomposition of $F$, and so there is a subset $\mathcal{S} \subseteq \mathcal{P}$ of size at least $|V|/4$ such that all potentials $P \in \mathcal{S}$ are pairwise disjoint, and every $P \subseteq V$. Let $Y = \cup_{P \in \mathcal{S}} P$ be the set of vertices that participate in this matching.

For each such $P = \{a, a'\} \in \mathcal{S}$, fix an arbitrary "default" assignment $\mathbf{x}_P^0$ for the vertices $\{a, a'\}$. Let $P^- \triangleq Y \setminus P$, and furthermore, let $\mathbf{x}_{P^-}^0$ be the assignment in which all variables in $P^-$ are set to their default assignments. For each $P = \{a, a'\} \in \mathcal{S}$, also let $\hat{X}_P = (X_a \times X_{a'}) \setminus \{\mathbf{x}_P^0\}$ be all possible assignments of $P$ that are not the default assignment. Finally, let $\mathbf{x}'_{V \setminus Y}$ be an arbitrary joint assignment for the remaining variables.

For each $P \in \mathcal{S}$, and each each $\hat{\mathbf{x}}_P \in \hat{X}_P$, consider the complete joint assignments of the form $complete(\hat{\mathbf{x}}_P) = (\hat{\mathbf{x}}_P, \mathbf{x}_{P^-}^0, \mathbf{x}'_{V \setminus Y})$. It's not hard to see that the coefficient vectors corresponding these assignments are all linearly independent, since $\mathbf{v}(complete(\hat{\mathbf{x}}_P))$ is the only such coefficient vector that places a 1 in the component corresponding to the potential $f_P$ and joint assignment $\hat{\mathbf{x}}_P$. Thus, $\mathbf{M}_F$ contains at least $|V|/4$ linearly independent columns, one for each of these complete assignments.

If the largest matching of $G_F$ does not contain at least $|V|/4$ edges, then it contains at most $|V|/2$ vertices. Thus $B \triangleq V \setminus Y$ contains at least $|A|/2$ vertices. There cannot be an edge between two vertices in $B$, otherwise there would have been a larger matching. Thus for any $a \in B$, and $P \in \mathcal{P}$ with $a \in P$, we have $P \cap B = \{a\}$. Since $G_F$ is an interaction graph, there must be at least one such $P$ for each $a$, call it $P_a$. As in the previous case, we can construct a linearly independent set of at least $|V|/2$ coefficient vectors by fixing all vertices to "default" values, considering each possible assignment of each variable $a \in B$, and finally considering the component of the coefficient vector corresponding to $P_a$ and the particular assignment of $a$. $\square$

**Example 1.** *There exists a class of functions $\mathcal{F}$ such that $\mathbf{M}_F$ has close to full rank for every $F \in \mathcal{F}$. In other words,* rank($\mathbf{M}_F$) = $\Omega(N)$, *where $\mathbf{M}_F \in \{0, 1\}^{N \times M}$ and $M > N$.*

*Proof.* Let $\mathcal{F}_n$ consist of functions $F$ which are the sum of unary potentials, so $F = \sum_{i=1}^n f_i(\mathbf{x}_i)$. Let $\mathbf{X} = \mathbf{X}_1 \times ... \times \mathbf{X}_n$ where each $\mathbf{X}_i = \{0, 1\}$. $M = 2^n$, while $N = 2n$. Let $\mathbf{x}^i$ be the joint assignment with $\mathbf{x}_i^i = 1$ and $\mathbf{x}_i^j = 0$ for all $j \neq i$. It's clear that $\{\mathbf{v}(\mathbf{x}_j^i)\}_j$ are linearly independent, establishing the claim. $\square$

**Example 2.** *Any function $F$ that does not decompose (i.e. with $\mathcal{P} = \{V\}$), has $\mathbf{M}_F$ equal to the identity matrix of size $|\mathbf{X}|$, and thus has* rank($\mathbf{M}_F$) = $\prod_{i \in V} |X_i|$.

## 9 Conclusions and Future Work

In this paper we described a new algorithm for MAB problems in which the unknown expected payoff function obeys structural constraints that are encoded in a graph. When the graph has bounded treewidth and bounded degree, our algorithm is tractable and suffers low regret. An important limitation of our approach is that the graphical structure of the expected payoff function must be specified in advance, and we would like to extend our algorithm so that it learns this structure as well. This will likely be challenging, since most formulations of structure learning in probabilistic graphical models are NP-hard. Our general approach will be to adapt tractable approximation algorithms for structure learning to our setting.

### Acknowledgments

We thank the anonymous reviewers for their helpful comments.